\documentclass{esannV2}
\usepackage{graphicx}
\usepackage[latin1]{inputenc}
\usepackage{amssymb,amsmath,array}
\usepackage{bm}
\usepackage{appendix}
\usepackage[square,sort,comma,numbers]{natbib}
\usepackage{subfig} 
\usepackage[hidelinks]{hyperref}
\usepackage[skip=3pt]{caption}
\newcommand{\ica}{\hspace{0.25cm}}

\renewcommand*{\bibfont}{\footnotesize}

\newcommand{\given}{\,|\,}

\begin{document}

\title{Sensitivity Analysis for Predictive Uncertainty in Bayesian Neural Networks}

\author{Stefan Depeweg$^{1,2}$,  Jos\'e Miguel Hern\'andez-Lobato$^3$,   Steffen Udluft$^2$,  Thomas Runkler$^{1,2}$
%
%
\vspace{.3cm}\\
%
1 - Technical Unversity of Munich, Germany. 2 - Siemens AG, Germany. \\
3 - University of Cambridge, United Kingdom.
}

\maketitle

\begin{abstract}
We derive a novel sensitivity analysis of input variables for predictive
epistemic and aleatoric uncertainty. We use Bayesian neural networks with
latent variables as a model class and illustrate the usefulness of our
sensitivity analysis on real-world datasets. Our method increases the
interpretability of complex black-box probabilistic models.
\end{abstract}
\vspace{-0.15cm}
\section{Introduction}
\vspace{-0.2cm}
Extracting human-understandable knowledge out of black-box machine learning
methods is a highly relevant topic of research. One aspect of this  is to
figure out how sensitive the model response is to which  input variables. This
can be useful both as a sanity check, if the  approximated function is reasonable,
but also to gain new insights about the problem at hand.  For neural
networks this kind of
model inspection can be performed by a sensitivity analysis \cite{fu1993sensitivity,montavon2017methods},
 a simple method that works  by considering
the gradient of the network output with respect to the input variables.  

Our key contribution is to transfer this idea towards predictive
uncertainty: What features impact the uncertainty in the predictions of our
model? To that end we use Bayesian neural networks (BNN) with latent variables
\cite{depeweg2016learning,depeweg2017decomposition}, a recently introduced
probabilistic model that can describe complex stochastic patterns while at the
same time account for model uncertainty. From their predictive distributions we
can extract epistemic and aleatoric uncertainties
\cite{kendall2017uncertainties,depeweg2017decomposition}. The former uncertainty
originates from our lack of knowledge of model parameter values and is
determined by the amount of available data, while aleatoric uncertainty consists of
irreducible stochasticity originating from unobserved (latent) variables. By
combining the sensitivity analysis with a decomposition of predictive
uncertainty into its epistemic and aleatoric components, we can analyze which
features influence each type of uncertainty. The
resulting sensitivities can provide useful insights into the model at hand. On
one hand, a feature with high epistemic sensitivity suggests that careful
monitoring or safety mechanisms are required to keep the values of this feature
in regions where the model is confident.  On the other hand, a feature with
high aleatoric uncertainty indicates a dependence of that feature with other
unobserved/latent  variables.
\vspace{-0.2cm}
\section{Bayesian Neural Networks with Latent Variables}
\vspace{-0.2cm}
Bayesian Neural Networks(BNNs) are scalable and flexible probabilistic models.
Given a training set~${\mathcal{D} = \{ \mathbf{x}_n, \mathbf{y}_n \}_{n=1}^N}$, formed
by feature vectors~${\mathbf{x}_n \in \mathbb{R}^D}$ and targets~${\mathbf{y}_n
\in \mathbb{R}}^K$, we assume that~${\mathbf{y}_n =
f(\mathbf{x}_n,z_n;\mathcal{W}) + \bm \epsilon_n}$, where~$f(\cdot ,
\cdot;\mathcal{W})$ is the output of a neural network with weights
$\mathcal{W}$ and $K$ output units. The network receives as input the feature vector $\mathbf{x}_n$ and the
latent variable $z_n \sim \mathcal{N}(0,\gamma)$.
We choose rectifiers:~${\varphi(x) = \max(x,0)}$ as activation functions for the hidden layers and
and the identity function:~${\varphi(x) = x}$ for the output layer.
The network output is
corrupted by the additive noise variable~$\bm \epsilon_n \sim \mathcal{N}(\bm 0,\bm
\Sigma)$ with diagonal covariance matrix $\bm \Sigma$. The role of the latent variable $z_n$ is to
capture unobserved stochastic features that can affect the network's
output in complex ways. 
The network has~$L$ layers, with~$V_l$ hidden units in layer~$l$,
and~${\mathcal{W} = \{ \mathbf{W}_l \}_{l=1}^L}$ is the collection of~${V_l
\times (V_{l-1}+1)}$ weight matrices.  
The $+1$ is introduced here to account
for the additional per-layer biases. 

We approximate the exact posterior distribution $p(\mathcal{W},\mathbf{z}\given\mathcal{D})$
with: 
\begin{align}
q(\mathcal{W},\mathbf{z}) =  & \underbrace{\left[ \prod_{l=1}^L\! \prod_{i=1}^{V_l}\!  \prod_{j=1}^{V_{l\!-\!1}\!+\!1} \mathcal{N}(w_{ij,l}| m^w_{ij,l},v^w_{ij,l})\right]}_{\text{\small $q(\mathcal{W})$}} \times \underbrace{\left[\prod_{n=1}^N \mathcal{N}(z_n \given m_n^z, v_n^z) \right]}_{\text{\small $q(\mathbf{z})$}}\,.
\label{eq:posterior_approximation}
\end{align}
The parameters~$m^w_{ij,l}$,~$v^w_{ij,l}$ and
~$m^z_n$,~$v^z_n$ are determined by
minimizing a divergence between 
$p(\mathcal{W},\mathbf{z}\given\mathcal{D})$
and the approximation $q$. For more detail the reader is referred to the work of 
\cite{depeweg2016learning,hernandez2016black}. In  our experiments we use
black-box $\alpha$-divergence minimization with $\alpha=1.0$.
\vspace{-0.2cm}
\subsection{Uncertainty Decomposition}\label{sec:unc_decomp}
\vspace{-0.2cm}
%
BNNs with latent variables can describe complex stochastic
patterns while at the same time account for model uncertainty. They achieve
this by jointly learning $q(\mathbf{z})$, which captures specific values of
the latent variables in the training data, and $q(\mathcal{W})$, which
represents any uncertainty about the model parameters. The result is a principled
Bayesian approach for learning flexible stochastic functions. 

For these models, we can
identify two types of uncertainty: \emph{aleatoric} and \emph{epistemic}
\citep{KIUREGHIAN2009105,kendall2017uncertainties}. Aleatoric
uncertainty originates from random latent
variables, whose randomness cannot be
reduced by collecting more data. In the BNNs this is given 
by $q(\mathbf{z})$ (and constant additive Gaussian noise $\bm \epsilon$, which we omit). Epistemic
uncertainty, on
the other hand, originates from lack of statistical evidence and
can be reduced by gathering more data. In the BNN this is given by
$q(\mathcal{W})$, which captures uncertainty over the model parameters.
These two forms of uncertainty are entangled in the approximate predictive distribution for a test input $\mathbf{x}^\star$:
\begin{equation}
p(\mathbf{y}^\star|\mathbf{x}^\star) = 
\int p(\mathbf{y}^\star|\mathcal{W},\mathbf{x}^\star,z)p(z^\star)q(\mathcal{W})\,dz^\star\,d\mathcal{W} \,.\label{eq:final_predictive_dist}
\end{equation}
where
$p(\mathbf{y}^\star|\mathcal{W},\mathbf{x}^\star,z^\star)=\mathcal{N}(\mathbf{y}^\star|f(\mathbf{x}^\star,z^\star;\mathcal{W}),\bm
\Sigma)$ is the likelihood function of the BNN and
$p(z^\star)=\mathcal{N}(z^\star|0,\gamma)$ is the prior on the latent
variables. 

We can use the variance $\sigma^2(y^\star_k|\mathbf{x}^\star)$ as a measure of predictive uncertainty for the $k$-th component of 
$\mathbf{y}^\star$. The variance can be decomposed into an epistemic and aleatoric term using the law of total variance:
\begin{align}
\sigma^2(y^\star_k|\mathbf{x}^\star)  &= \sigma^2_{q(\mathcal{W})}(\mathbf{E}_{p(z^\star)}[y^\star_k|\mathcal{W},\mathbf{x}^\star])
+ \mathbf{E}_{q(\mathcal{W})}[\sigma^2_{p(z^\star)}(y^\star_k|\mathcal{W},\mathbf{x}^\star)]
\label{eq:decomp}
\end{align}
The first term, that is 
$\sigma^2_{q(\mathcal{W})}(\mathbf{E}_{p(z^\star)}[y^\star_k|\mathcal{W},\mathbf{x}^\star])$ is the
variability of $y^\star_k$, when we integrate out $z^\star$ but not $\mathcal{W}$.
Because $q(\mathcal{W})$ represents our belief over model parameters, this is  a measure of the $ \emph{epistemic}$ uncertainty. The second term, 
$\mathbf{E}_{q(\mathcal{W})}[\sigma^2_{p(z^\star)}(y^\star_k|\mathcal{W},\mathbf{x}^\star)]$ represents the average variability of 
$y^\star_k$ not
originating from the distribution over model parameters $\mathcal{W}$. This measures \emph{aleatoric} uncertainty, as the variability can only come from the latent variable $z^\star$.
\vspace{-0.2cm}
\section{Sensitivity Analysis of Predictive Uncertainty}
\vspace{-0.2cm}
In this section we will extend the method of sensitivity analysis toward predictive uncertainty. The goal is to
provide insight into the question of which features affect the  stochasticity of our model, which results in aleatoric uncertainty,  and which
features impact its epistemic uncertainty.  For instance, if we have limited data about different settings of a particular feature $i$, even a small change of its value can have a large effect on the confidence of the model. 

Answers to these two questions can provide useful insights about a model at
hand. For instance, a feature with high aleatoric sensitivity indicates a
strong interaction with other unobserved/latent features. If a practitioner can
expand the set of features by taking more refined measurements, it may be
advisable to look into variables which may exhibit dependence with that feature
and which may explain the stochasticity in the data. Furthermore, a feature
with high epistemic sensitivity, suggests careful monitoring or extended safety
mechanisms are required to keep this feature values in regions where the model
is confident.

We start by briefly reviewing the technique of sensitivity analysis \cite{fu1993sensitivity,montavon2017methods}, a simple  method that can provides insight into how changes in the input affect the network's prediction.
 Let $\mathbf{y} = f(\mathbf{x};\mathcal{W})$ 
be a neural network fitted on a training set~${\mathcal{D} = \{ \mathbf{x}_n, \mathbf{y}_n \}_{n=1}^N}$, formed
by feature vectors~${\mathbf{x}_n \in \mathbb{R}^D}$ and targets~${\mathbf{y}_n
\in \mathbb{R}}^K$.  We want to understand how each feature $i$ influences the output dimension $k$. 
Given some test data $\mathcal{D}_\text{test}=\{ \mathbf{x}_n^\star, \mathbf{y}_n^\star \}_{n=1}^{N_\text{test}}$, we use the partial derivate of the output dimension $k$ w.r.t. feature $i$:
\begin{equation}
I_{i,k} =   \frac{1}{N_\text{test}} \sum_{n=1}^{N_\text{test}}  \big| \frac{\partial f(\mathbf{x}^\star_n)_k}{\partial x^\star_{i,n}} \big|\,.
\label{eq:sensitivity}
\end{equation}

In Section \ref{sec:unc_decomp} we saw that we can decompose the variance of the predictive distribution of a BNN with latent variables into its epistemic and aleatoric components. Our goal is to obtain sensitivities
of these components with respect to the input variables.  For this we use  a sampling based approach to approximate
 the two uncertainty components \cite{depeweg2017decomposition} and  then calculate the partial derivative of these w.r.t. to the input variables. For each test data point $\mathbf{x}_n^\star$, we perform $N_w \times N_z$
forward passes through the BNN. We first sample $w \sim q(\mathcal{W})$ a total of $N_w$ times and
then, for each of these samples of $q(\mathcal{W})$, performing $N_z$ forward passes in
which $w$ is fixed and we only sample the latent variable $z$.  Then we can do an empirical estimation of the expected predictive value and of the two components on the right-hand-side of Eq. (\ref{eq:decomp}):
\begin{align}
\mathbf{E}[y^\star_{n,k}|\mathbf{x}_n^\star] &\approx \frac{1}{N_w} \frac{1}{N_z} \sum_{n_w=1}^{N_w} \sum_{n_z=1}^{N_z}  y^\star_{n_w,n_z}(\mathbf{x}^\star_n)_k \\
\sigma_{q(\mathcal{W})}(\mathbf{E}_{p(z^\star)}[y^\star_{n,k}|\mathcal{W},\mathbf{x}^\star_n]) &\approx \hat{\sigma}_{N_w}( \frac{1}{N_z}\sum_{n_z=1}^{N_z} y^\star_{n_w,n_z}(\mathbf{x}^\star_n)_k) \\
\mathbf{E}_{q(\mathcal{W})}[\sigma^2_{p(z^\star)}(y^\star_{n,k}|\mathcal{W},\mathbf{x}^\star_n)]^{\frac{1}{2}} &\approx \big(\frac{1}{N_w}\sum_{n_w=1}^{N_w} \hat{\sigma}^2_{N_z}(y^\star_{n_w,n_z}(\mathbf{x}^\star_n)_k)\big)^{\frac{1}{2}} \,.
\end{align}
where $y^\star_{n_w,n_z}(\mathbf{x}^\star_n)_k = f(\mathbf{x}^\star_n,z^{n_w,n_z};\mathcal{W}^{n_w})_k$
and $\hat{\sigma}^2_{N_z}$ ($\hat{\sigma}^2_{N_w}$) is an empirical estimate of the variance over $N_z$ ($N_w$) samples of $z$ ($\mathcal{W}$). We have used the square root of each component so all terms share the same unit of $y^\star_{n,k}$. Now we can calculate  the sensitivities:
\begin{align}
I_{i,k} &=  \frac{1}{N_\text{test}} \sum_{n=1}^{N_\text{test}}  \big| \frac{\partial \mathbf{E}[y^\star_{n,k}|\mathbf{x}^\star_n])}{\partial x^\star_{i,n}} \big| \label{eq:expected_sensitivity} \\
I_{i,k}^{\text{epistemic}} &=   \frac{1}{N_\text{test}} \sum_{n=1}^{N_\text{test}} \big|\frac{\partial \sigma_{q(\mathcal{W})}(\mathbf{E}_{p(z^\star)}[y^\star_{n,k}|\mathcal{W},\mathbf{x}^\star_n])}{\partial x^\star_{i,n}}\big| \label{eq:epistemic_sensitivity} \\
I_{i,k}^{\text{aleatoric}} &=   \frac{1}{N_\text{test}} \sum_{n=1}^{N_\text{test}} \big|\frac{\partial \mathbf{E}_{q(\mathcal{W})}[\sigma^2_{p(z^\star)}(y^\star_{n,k}|\mathcal{W},\mathbf{x}^\star_n)]^{\frac{1}{2}}}{\partial x^\star_{i,n}}\big| \label{eq:aleatoric_sensitivity} \,, 
\end{align}
where Eq. (\ref{eq:expected_sensitivity}) is the standard sensitivity term. We also note that the general drawbacks \cite{montavon2017methods}  of the sensitivity analysis,  such as considering every variable in isolation,  arise due to its simplicity. These will also apply when focussing on the uncertainty components.  
\vspace{-0.25cm}
\section{Experiments}
\vspace{-0.2cm}
\begin{figure}[h!]
\centering
\subfloat[][]{\includegraphics[width=0.30\linewidth]{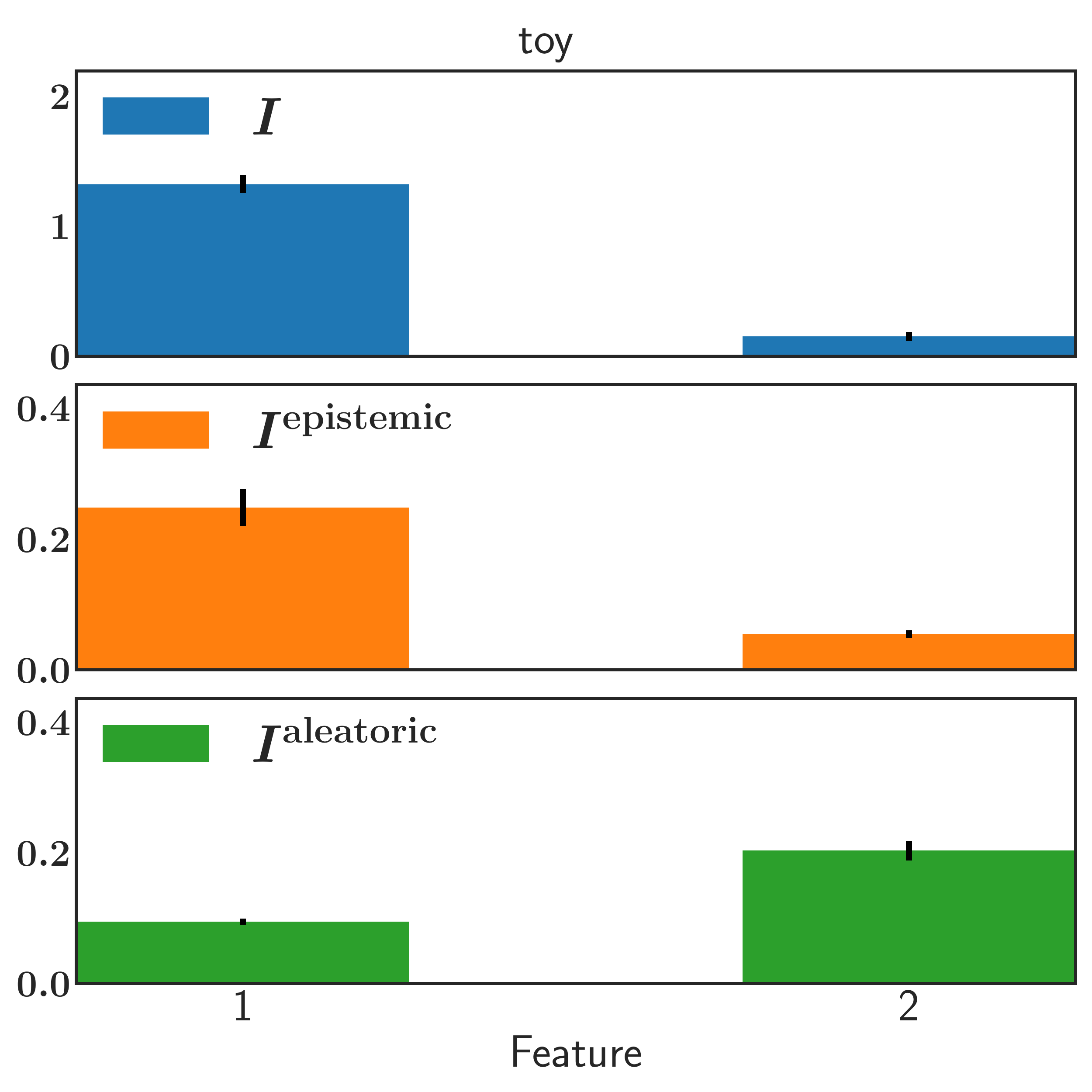}\label{fig:toy}}
\subfloat[][]{\includegraphics[width=0.30\linewidth]{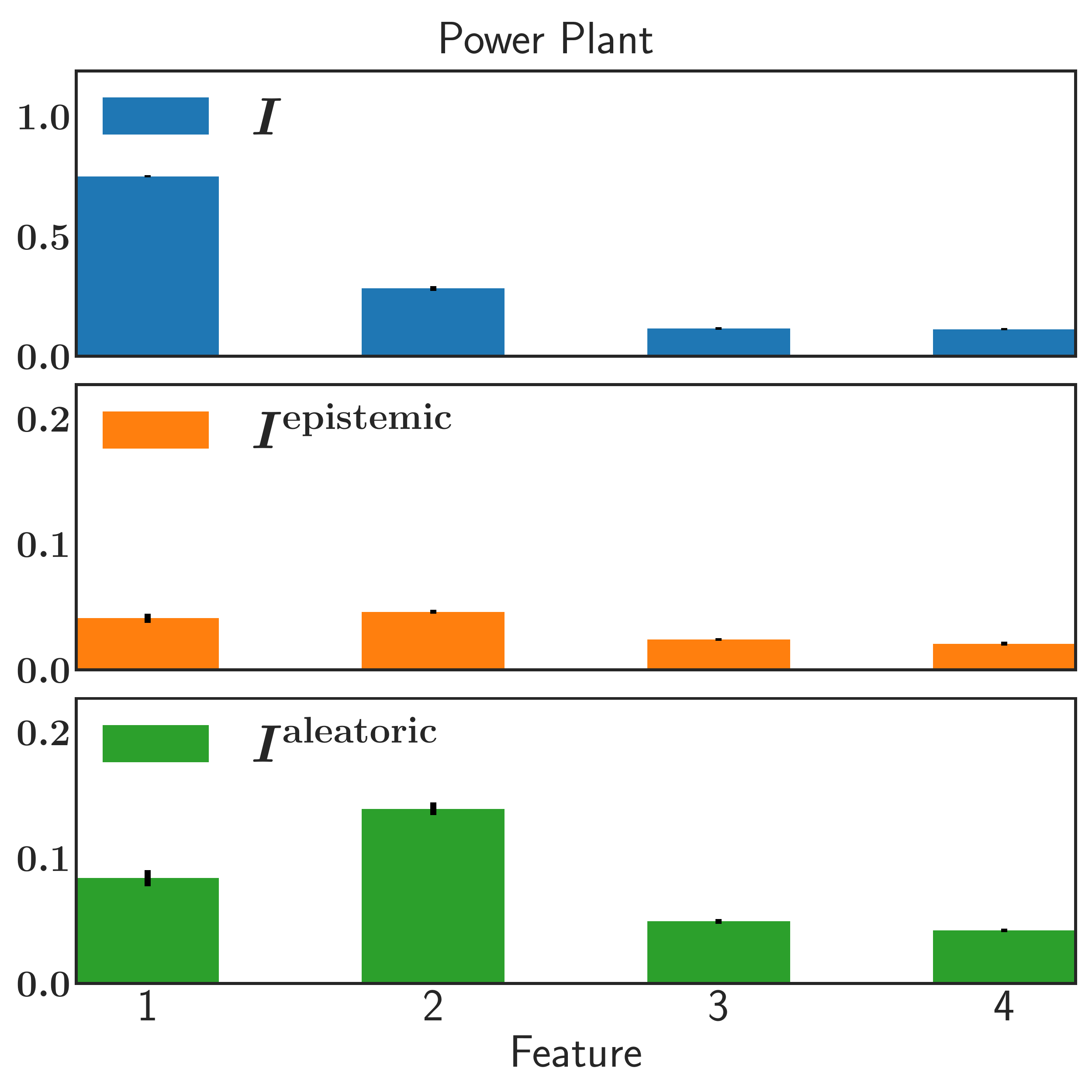}\label{fig:pplant}}
\subfloat[][]{\includegraphics[width=0.30\linewidth]{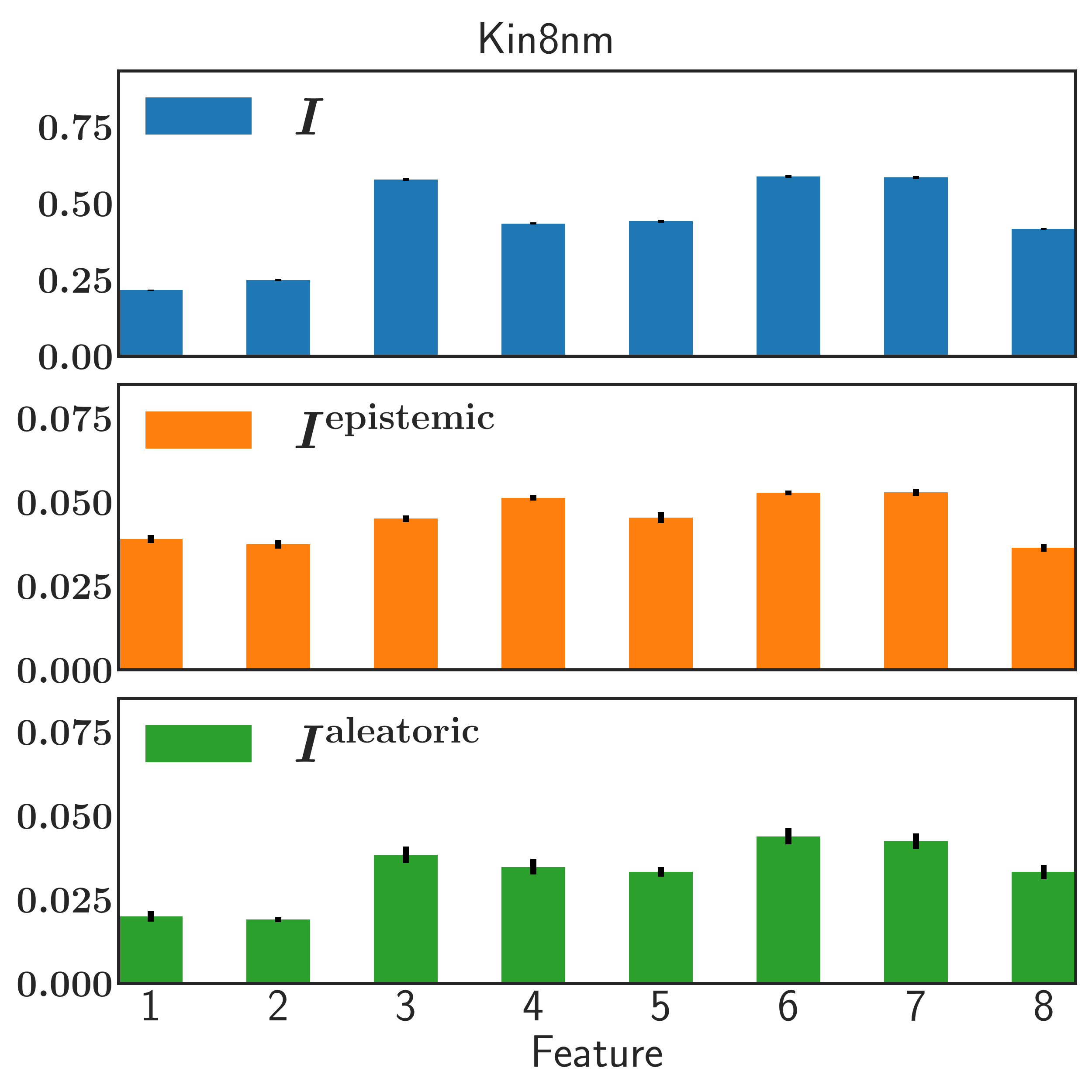}\label{fig:kin8nm}} \\
\vspace{-0.3cm}
\subfloat[][]{\includegraphics[width=0.30\linewidth]{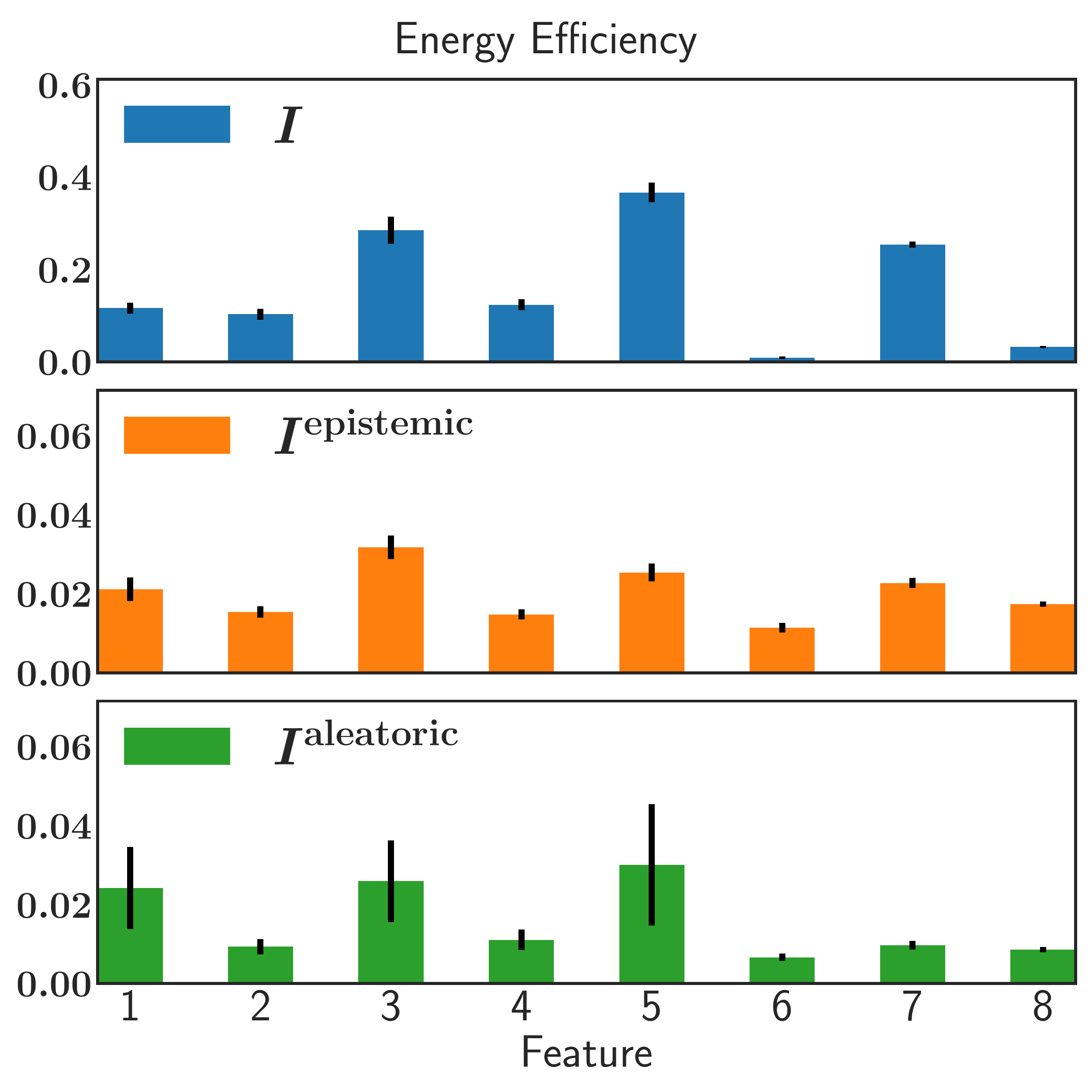}\label{fig:energy}}
\subfloat[][]{\includegraphics[width=0.30\linewidth]{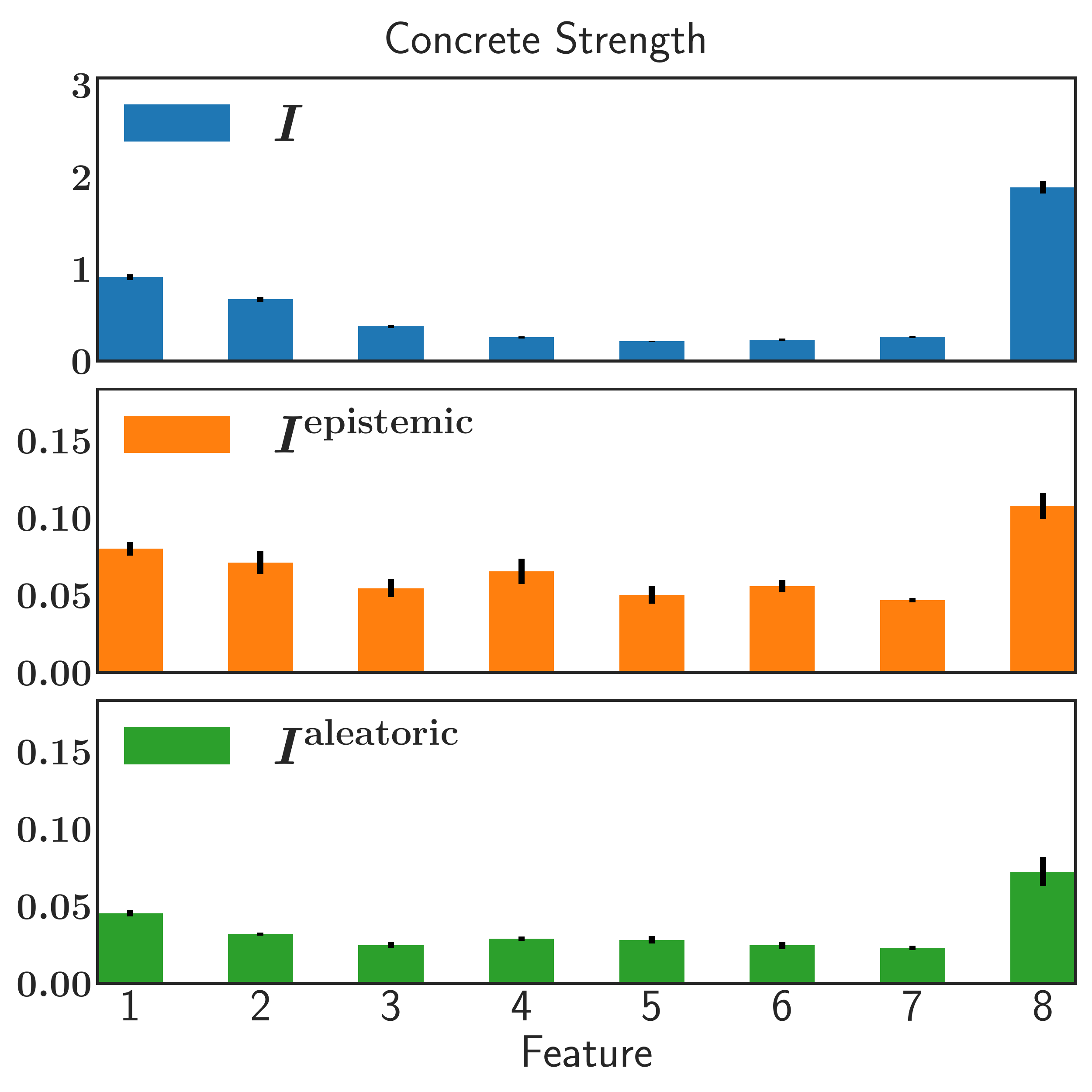}\label{fig:concrete}}
\subfloat[][]{\includegraphics[width=0.30\linewidth]{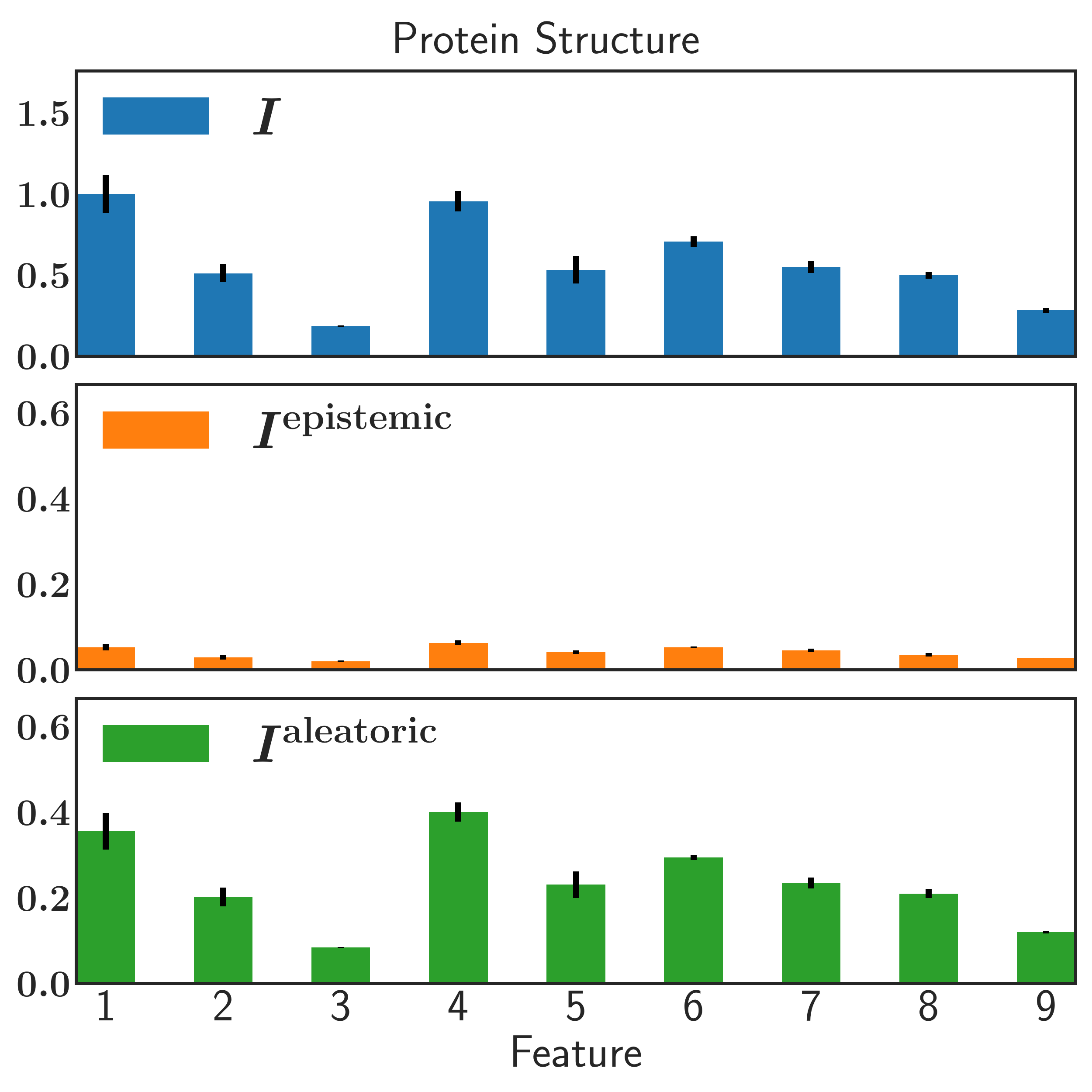}\label{fig:protein}}  \\ 
\vspace{-0.3cm}
\subfloat[][]{\includegraphics[width=0.30\linewidth]{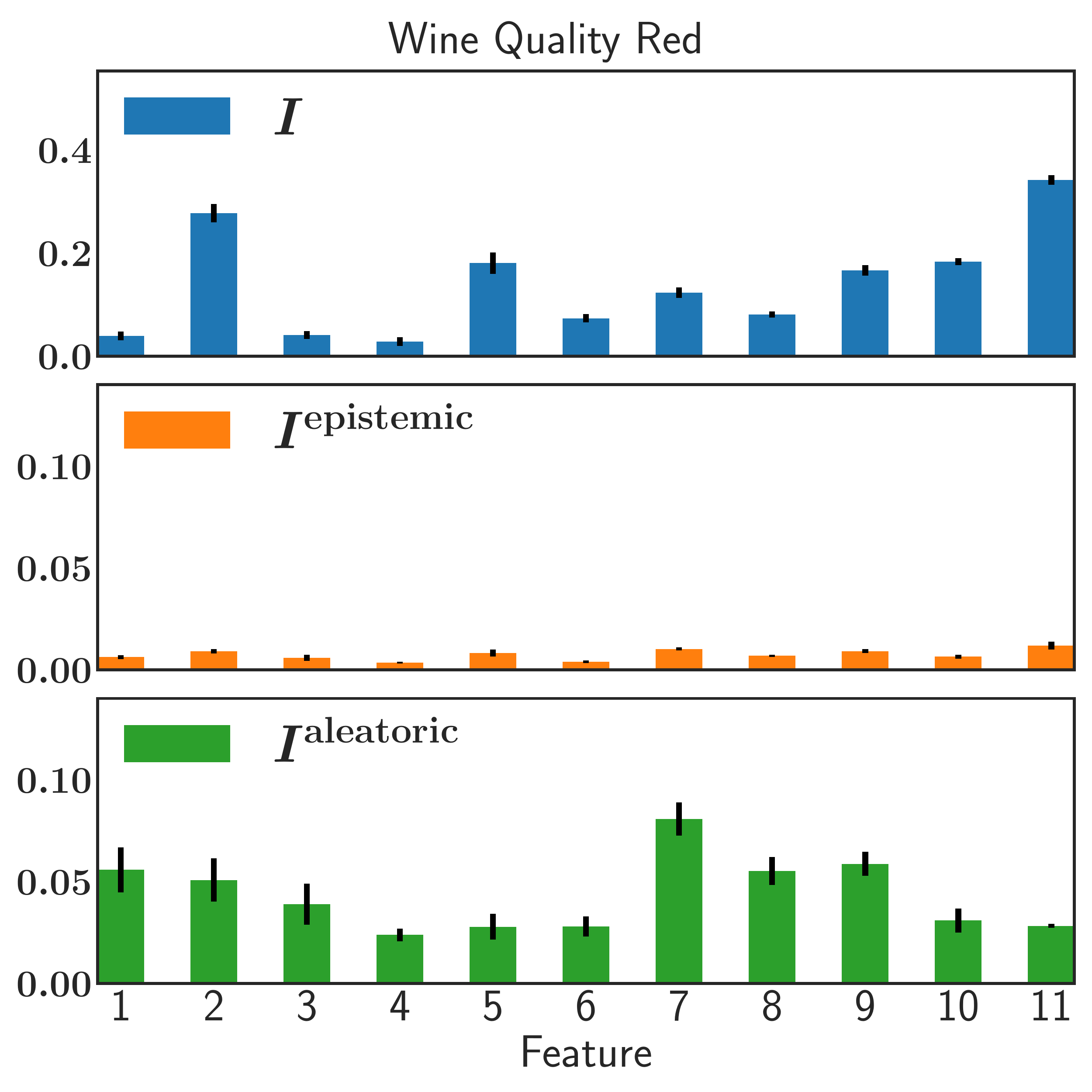}\label{fig:wine}}
\subfloat[][]{\includegraphics[width=0.30\linewidth]{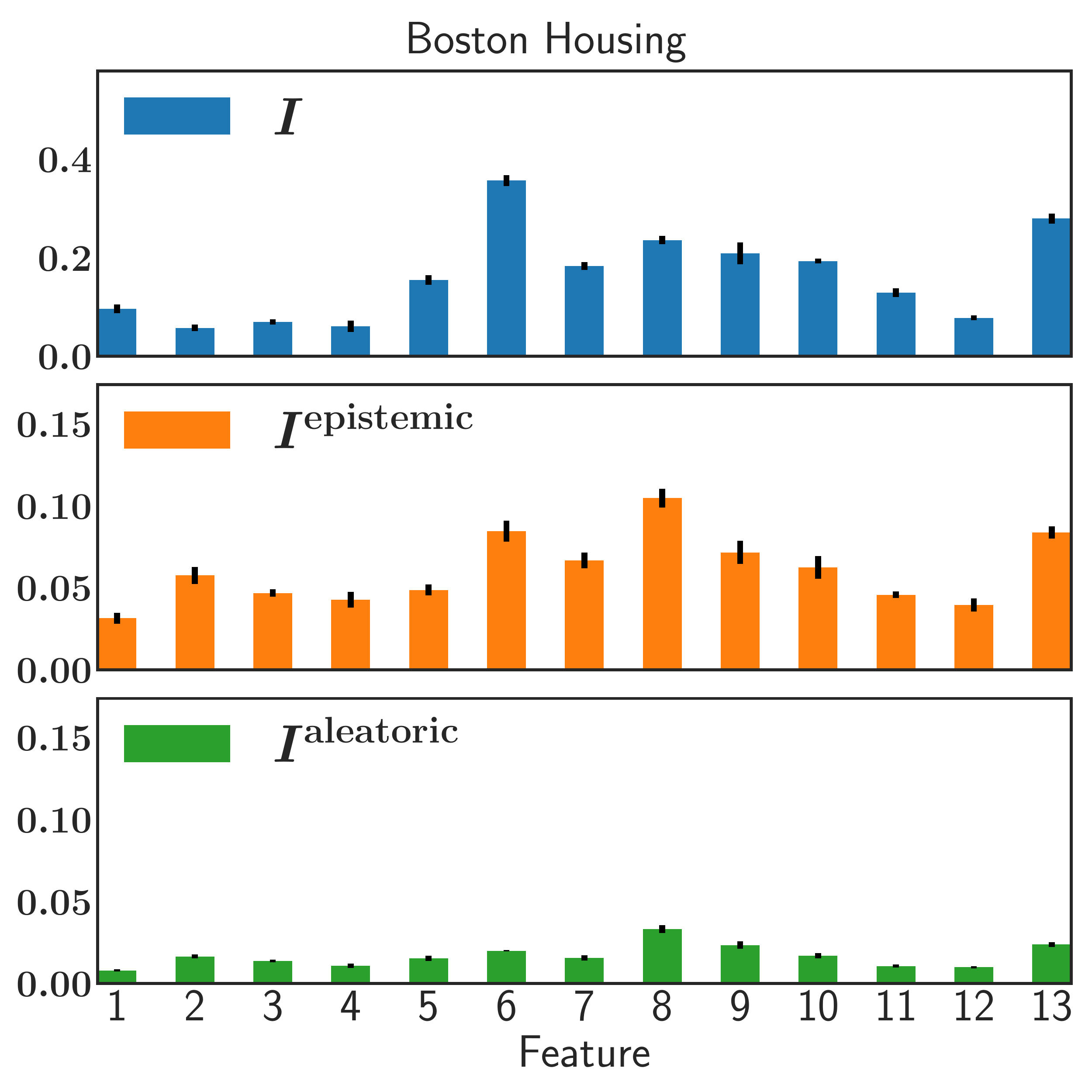}\label{fig:boston}} 
\subfloat[][]{\includegraphics[width=0.30\linewidth]{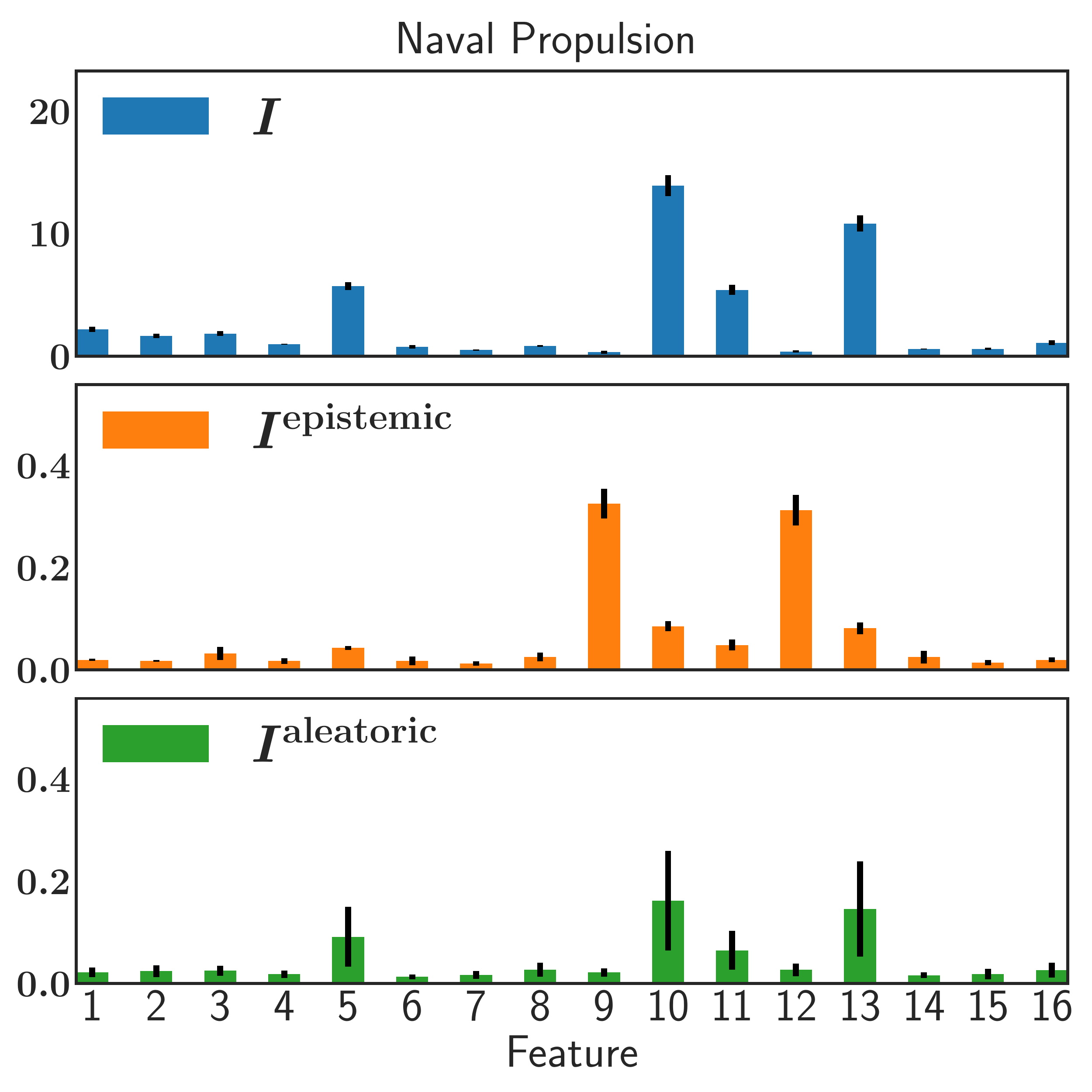}\label{fig:naval}} 
\caption{Sensitivity analysis for the predictive expectation and uncertainty on toy data \protect\subref{fig:toy} and UCI datasets \protect\subref{fig:pplant}-\protect\subref{fig:naval}. Top row shows  sensitivities w.r.t. expectation (Eq. (\protect\ref{eq:expected_sensitivity})). Middle and bottom row show  sensitivities for epistemic and aleatoric uncertainty (Eq.~(\protect\ref{eq:epistemic_sensitivity}) and Eq.~(\protect\ref{eq:aleatoric_sensitivity})). Error bars indicate standard errors over experiment repetitions.}
\label{fig:uci}
\vspace{-0.25cm}
\end{figure}
In this section we want to do an exploratory study. For that we will first use an artifical toy dataset and then use 8 datasets from the UCI repository \cite{lichman2013uci} in varying domains and dataset sizes.  For all experiments, we use a BNN with 2 hidden layer. We first perform model selection on the number of hidden units per layer from $\{20,40,60,80\}$  on the available data, details can be found in Appendix \ref{ap:ms}. We train for $3000$ epochs with a learning rate of $0.001$ using Adam as optimizer. For the sensitivity analysis we will sample $N_w=200$ $w \sim q(\mathcal{W})$ and and $N_z=200$ samples from $z \sim \mathcal{N}(0,\gamma)$. All experiments were repeated $5$ times and we report average results. 
\vspace{-0.15cm}
\subsection{Toy Data}
\vspace{-0.15cm}
We consider a regression task for a stochastic function with heteroskedastic
noise: $y= 7 \sin (x_1) + 3|\cos (x_2 / 2)| \epsilon$ with $\epsilon \sim
\mathcal{N}(0,1)$.  The first input variable $x_1$ is responsible for the shape
of the function whereas the second variable $x_2$ determines the noise level.  We sample $500$ data points
with $x_1 \sim \text{exponential}(\lambda=0.5)-4$ and $x_2 \sim \mathcal{U}(-4,4)$. Fig. \ref{fig:toy} shows the sensitivities. The first variable $x_1$ is responsible for the epistemic uncertainty whereas $x_2$ is responsible for the aleatoric uncertainty which corresponds 
with the generative model for the data.
\vspace{-0.15cm}
\subsection{UCI Datasets}
\vspace{-0.15cm}
We consider several real-world regression datasets from the UCI data repository \cite{lichman2013uci}. Detailed
descriptions, including feature and target explanations, can be found on the respective
website. For evaluation we use the same training and test data splits as in \cite{hernandez2016black}. 
In Fig. \ref{fig:uci} we show the results of all experiments. For some problems the aleatoric sensitivity is most prominent (Fig. \ref{fig:protein},\ref{fig:wine}), while in others we have predominately epistemic sensitivity (Fig. \ref{fig:concrete},\ref{fig:boston}) and a mixture in others. This makes sense, because we have variable dataset sizes (e.g. Boston Housing with 506 data points and 13 features, compared to Protein Structure with 45730 points and 9 features) and also likely different heterogeneity in the datasets.

In the power-plant example feature $1$ (temperature) and $2$ (ambient pressure) are the main sources of aleatoric uncertainty of the  target, the net hourly electrical energy output. The data in this problems originates
from a combined cycle power plant consisting of gas and steam turbines. The provided features likely provide only
limited information of the energy output, which is subject to complex combustion processes. We can expect that a change in temperature and pressure will influence this process in a complex way, which can explain the high sensitivities we see.
The task in the naval-propulsion-plant example, shown in Fig. \ref{fig:naval}, is to predict
the compressor decay state coefficient, of a gas turbine operated on a naval vessel. Here we see that two features, the compressor inlet air temperature and air pressure have high epistemic uncertainty, but do not influence the overall sensitivity much. This makes sense, because we only have a single value of both features in the complete dataset. The model has learned no influence of this feature on the output (because it is constant) but any change from this constant will make the system highly uncertain.
\vspace{-0.30cm}
\section{Conclusion}
\vspace{-0.15cm}
In this paper we provided a new way of sensitivity analysis for predictive epistemic and aleatoric uncertainty. Experiments indicate useful insights of this method on real-world datasets. 
\bibliographystyle{unsrt}
\vspace{-0.25cm}
{\small
\bibliography{references}
}

\clearpage
\begin{appendices}
\section{Model Selection}\label{ap:ms}
We perform model selection on the number of hidden units for a BNN with 2 hidden layer. Table \ref{tab:results_tll} shows test log-likelihoods with standard error on all UCI datasets.  By that we want to lower the effect 
of over- and underfitting. Underfitting would show itself by high aleatoric and low epistemic uncertainty whereas
overfitting would result in high epistemic and low aleatoric uncertainty.
 
\begin{table*}[!ht]
\centering
\caption{Test Log-likelihood on UCI Dataset.}
\begin{tabular}{l@{\ica}r@{$\pm$}l@{\ica}r@{$\pm$}l@{\ica}r@{$\pm$}l@{\ica}r@{$\pm$}l@{\ica}}\hline
&\multicolumn{8}{c}{Hidden units per layer} \\
\bf{Dataset}&\multicolumn{2}{c}{\bf{$20$}}&\multicolumn{2}{c}{\bf{ $40$ }}&\multicolumn{2}{c}{\bf{$60$}}&\multicolumn{2}{c}{\bf{ $80$ }}\\
\hline
{\bf Power Plant }  & -2.76&0.03&-2.71&0.01&{\bf -2.69}&{\bf 0.02}&-2.71&0.03 \\
{\bf Kin8nm} & 1.23&0.01&1.30&0.01&1.30&0.01&{\bf 1.31}&{\bf 0.01} \\
{\bf Energy Efficiency}&-0.79&0.10&{\bf -0.66}&{\bf 0.07}&-0.79&0.07&-0.74&0.06 \\
{\bf Concrete Strength} & -3.02&0.05&{\bf -2.99}&{\bf 0.03}&-3.00&0.01&-3.01&0.04 \\
{\bf Protein Structure} & -2.40&0.01&-2.32&0.00&-2.26&0.01&{\bf -2.24}&{\bf 0.01} \\
{\bf Wine Quality Red} &  1.94& 0.11&{\bf 1.95}&{\bf 0.09}&1.78&0.13&1.62&0.13 \\ 
{\bf Boston Housing} & -2.49&0.12&{\bf -2.39}&{\bf 0.05}&-2.52&0.03&-2.66&0.03 \\ 
{\bf Naval Propulsion} & 1.80&0.04&1.74&0.06&1.81&0.05&{\bf 1.86}&{\bf 0.05} \\
\hline
\end{tabular}
\label{tab:results_tll}
\end{table*}
\end{appendices}


\end{document}